\documentclass[conference]{IEEEtran}
\usepackage{times}

\usepackage[numbers]{natbib}
\usepackage{multicol}
\usepackage[bookmarks=true]{hyperref}
\usepackage{amsmath}
\usepackage{graphicx}
\usepackage{amssymb}
\usepackage{booktabs}%
\usepackage{caption}
\begin{document}

% paper title
\title{A spatiotemporal model with visual attention for video classification}

\author{
\authorblockN{Mo Shan and Nikolay Atanasov}
\authorblockA{Department of Electrical and Computer Engineering\\
University of California San Diego, La Jolla, California, USA\\
Email:  $\{$moshan, natanasov$\}$@eng.ucsd.edu}
}

\maketitle

\begin{abstract}

High level understanding of sequential visual input is important for safe and stable autonomy, especially in localization and object detection. 
While traditional object classification and tracking approaches are specifically designed to handle variations in rotation and scale, current state-of-the-art approaches based on deep learning achieve better performance. 
This paper focuses on developing a spatiotemporal model to handle videos containing moving objects with rotation and scale changes. 
Built on models that combine Convolutional Neural Networks (CNNs) and Recurrent Neural Networks (RNNs) to classify sequential data, this work investigates the effectiveness of incorporating attention modules in the CNN stage for video classification. The superiority of the proposed spatiotemporal model is demonstrated on the Moving MNIST dataset augmented with rotation and scaling. 

\end{abstract}

\IEEEpeerreviewmaketitle

\section{Introduction}

Semantic understanding of sequential visual input is vital for autonomous robots to perform localization and object detection. Moreover, robust models have to be adaptive to scenes containing multiple interacting objects. 
For instance, self-driving cars need to classify scenes that contains small and large vehicles for loop closure. Similarly, robots have to infer the classes of objects in order to grasp them, while the objects could be rotated in different ways.    
Although videos contain richer information than individual images, how to effectively harness the temporal dependency in sequential visual input remains an open problem. Particularly, this paper investigates video classification in the presence of moving objects with rotation and scale changes. 
Preliminary results on digit gesture classification demonstrate the extensibility of the proposed model on handling articulated objects such as hand. 

$\mathbf{Problem\ Formulation}$. Given a training dataset that contains videos $D_{train} = \{ (x_1, y_1), (x_2, y_2), ..., (x_n, y_n) \}$, where $x_i \in \mathbb{R}_{>0}^{w\times h\times c\times t}$ is the $i_{th}$ video, $w, h, c$ is the size for each frame, $t$ is the number of frames, and $y_i \in \{ 1, 2, ..., K \}$ is the $i_{th}$  class label, we aim to train a classifier $\mathbf{C}$ for unseen data $(x_i, y_i) \in D_{test}$, such that $\mathbf{C} (x_i) = y_i$. 
For instance, in the context of action recognition, each video $x_i$ is labeled with a class label $y_i$ such as walking or running. 

\begin{figure}[thpb]
  \centering
  \includegraphics[scale=0.45]{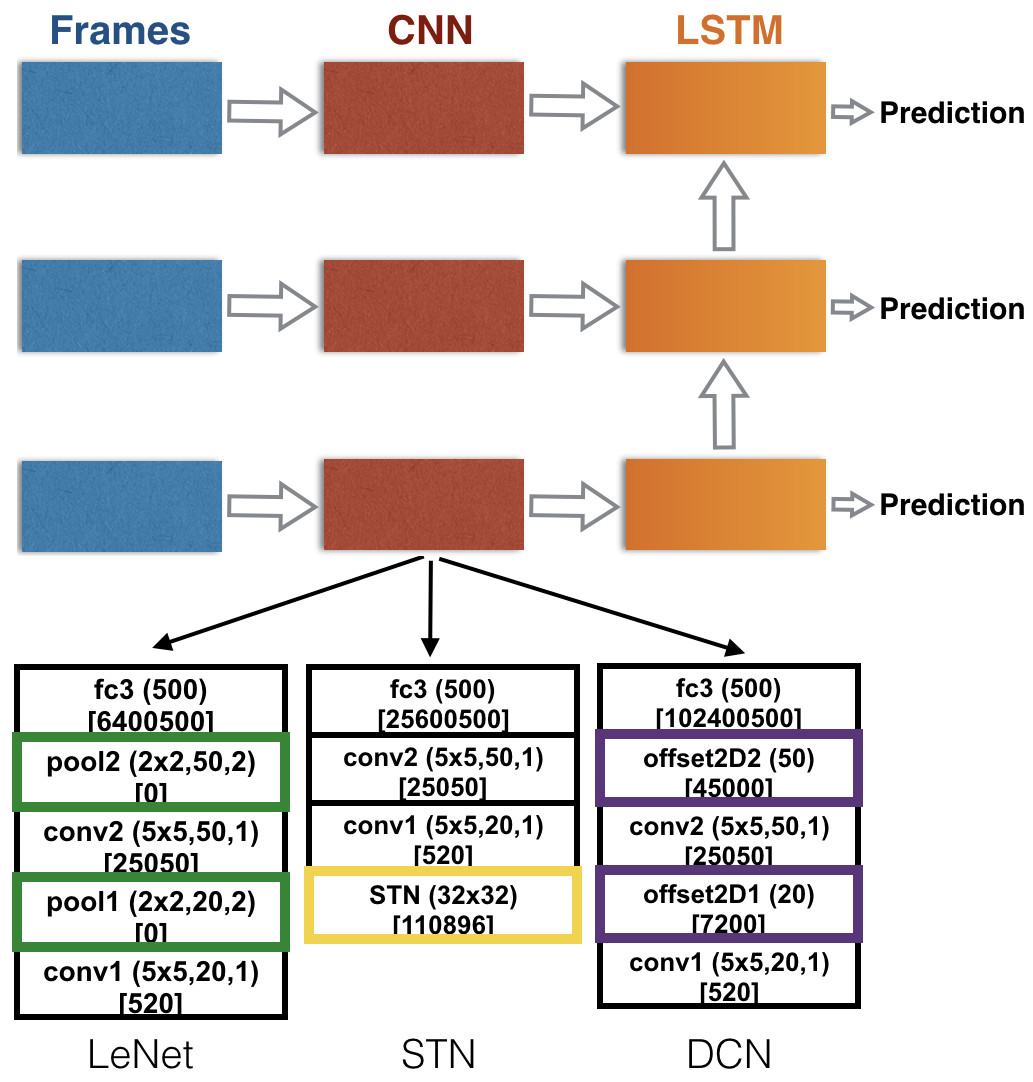}
  \caption{Architecture of the proposed model: The spatiotemporal model contains a CNN for feature extraction and a LSTM for temporal processing. Three CNN models are illustrated in the figure, including the baseline LeNet, and two variants of LeNet augmented with attention modules, one with STN and the other with DCN. The layers that handle geometric transformations are highlighted in the colored boxes.    
For the CNNs, conv is convolution layer, fc is fully-connected layer, pool is max-pooling layer. 
The numbers inside parentheses are: kernel size, number of output channels, and stride for conv and pool layers; number of output channels for offset2D layers; number of outputs for fc layers; size of output feature map for STN.
The numbers inside the square brackets are trainable parameters in each layer given a $64\times 64$ input image.}
  \label{fig:cnn_lstm}
\end{figure}

Initial work \cite{karpathy2014large} approaches this task using CNNs, but the majority of recent works such as \cite{sharma2015action} rely more on RNNs that are adept at processing sequential data.  
While RNNs are good at dealing with contextual information from the past, CNNs are useful to extract hierarchical features from video frames. Motivated by the advantages of both types of networks, this paper proposes a spatiotemporal model in which CNN and RNN are concatenated, as shown in Fig. \ref{fig:cnn_lstm}. 

Despite the impressive performance of deep neural networks, previous models have fixed network depth and convolutional kernel size, which may lead to brittle performance when dealing with large and rapid variations in scale and rotation of objects in high speed robot navigation.  
In the proposed spatiotemporal model, an attention module is introduced in the network architecture. This is the key idea that increases the robustness of the model to geometric transformations such as object rotation and scale changes commonly seen in the real world. 
Previously, networks with visual attention were developed to limit the amount of information that needs to be processed and increase their robustness against cluttered background. However, these attention modules have been only used for single image recognition and classification tasks. This work applies attention mechanism for video classification. 

Attention modules can be categorized into hard attention \cite{mnih2014recurrent} and soft attention \cite{jaderberg2015spatial, lin2016inverse, dai2017deformable}. This work focuses on the latter since the former is stochastic and non-differentiable. The soft attention modules considered in the proposed spatiotemporal model include Spatial Transformer Networks (STN) \cite{jaderberg2015spatial} and Deformable Convolutional Networks (DCN) \cite{dai2017deformable}. STN performs affine transformation on the feature map globally, whereas DCN samples the feature map locally and densely via learnt offsets to deform the convolutional kernel. LeNet \cite{lecun2015lenet} which has a simple structure and no attention module is used as baseline for comparison. 

To summarize, the novelty of the paper is a spatiotemporal model with visual attention tailored for video classification, which enables robustness to multiple objects with rotation and scale changes. 

\section{Spatial temporal model with visual attention}

\subsection{Feature extraction}

Given a sequence of images, hierarchical features are extracted using a model similar to LeNet. The lower layers of LeNet include two convolution and max-pooling blocks. Its upper layers contain a fully-connected layer, and a softmax layer used for classification during pre-training, but removed for feature extraction. Although the model is simple, it still contains the essence of deeper CNNs. The trainable parameters are the weights and bias terms in the convolutional and fully-connected layers shown in Fig. \ref{fig:cnn_lstm}. 
The proposed model uses an attention module prior to the first convolution layer instead of max-pooling layers to deal with geometric transformation, which means max-pooling layers are removed and the remaining CNN layers are augmented by attention modules. There are two variations of the structure, one with STN and the other with DCN. 

\begin{figure}[thpb]
  \centering
  \includegraphics[scale=0.43]{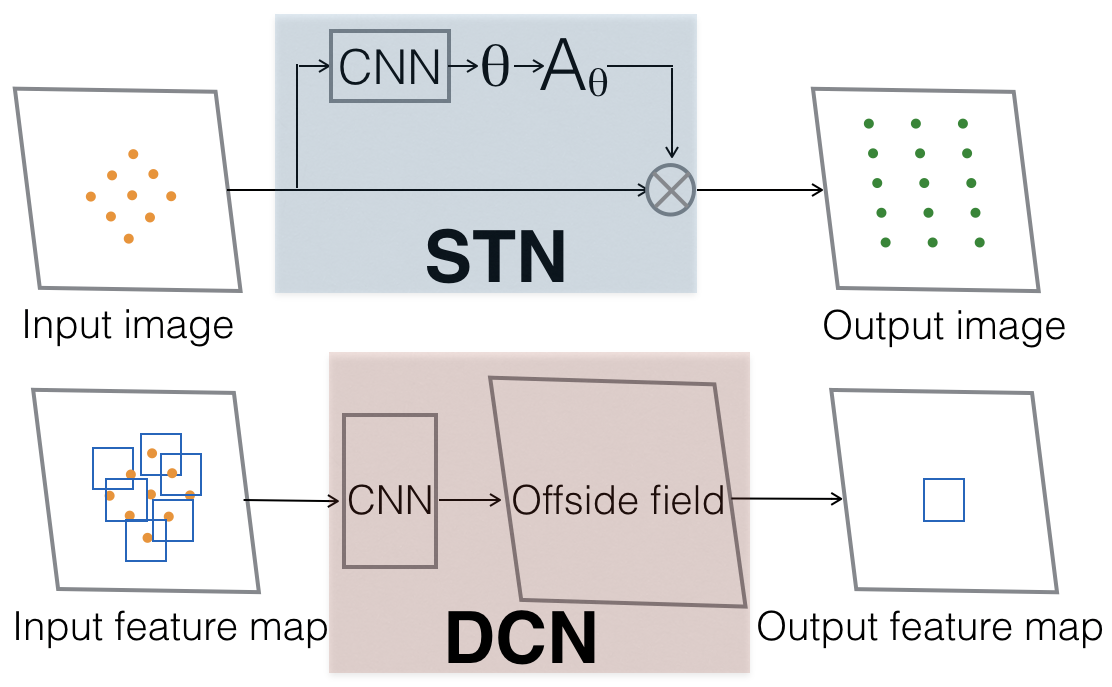}
  \caption{Structure of STN and DCN.}
  \label{fig:stn_dcn}
\end{figure}

STN consists of three components: localization network, grid generator, and bilinear sampler as depicted in Fig. \ref{fig:stn_dcn}. 
The localization network consists of 2 max-pooling layers, 2 convolution layers, and 2 fully-connected layers. It takes the image $\mathcal{I}$ as input and learns the affine transformation $A_{\theta}$ that should be applied to the image, i.e. $\theta = f_{loc}(\mathcal{I})$, where $\theta$ represents the transformation parameters.
The grid generator produces coordinates $A_{\theta} (\mathcal{G})$ based on $A_{\theta}$ and a regular grid $\mathcal{G}$, which are the sampling locations in the input image to produce the output feature map. 
The bilinear sampler takes input image and $A_{\theta} (\mathcal{G})$ to generate the output feature map using bilinear interpolation.  
Because of the affine transformation, STN is trained to focus on the digits via cropping, translation and isotropic scaling  of the input image. The output size of STN is $32\times 32$. Different output sizes of STN are tested, but that does not affect the classification accuracy significantly. 
Since additional layers are introduced in the localization network, there are more trainable parameters in the proposed model as described in Fig. \ref{fig:cnn_lstm}. 

DCN abandons the parametric transformation adopted by STN and its variants. As shown in Fig. \ref{fig:stn_dcn}, the spatial sampling locations in the convolution kernels are augmented with additional offsets learnt from classification. For example, a $5\times 5$ kernel with dilation 1 in a convolutional layer samples on grid $\mathcal{R} = \{(-2, -2), (-2, -1), ..., (1, 2), (2, 2)\}$ on input feature map $x$.The feature at $p_0$ in output feature map $y$ could be obtained based on Eq. (\ref{eq:dcn1}). The grid $\mathcal{R}$ is augmented by offsets $\{ \Delta p_n | n = 1, ..., N \}, N = |R|$ in deformable convolution as in Eq. (\ref{eq:dcn2}), where the offset is implemented by bilinear interpolation. $\Delta p_n$ is learnt via applying a convolutional layer over $x$, which has the same kernel size of $5\times 5$, and same spatial resolution with $x$. The output of the offset fields has $2N$ channels, which encodes $N$ vectors for $\Delta p_n$ in $2D$ denoted as offset2D layers in Fig. \ref{fig:cnn_lstm}. The trainable parameters include the offsets $\{ \Delta p_n | n = 1, ..., N \}$ as well as the convolutional kernels.  

\begin{subequations}
\label{eq:dcn}
\begin{align}
y(p_0) = \sum_{p_n \in \mathcal{R}} w(p_n) \cdot x(p_0 + p_n) \label{eq:dcn1}  \\
y(p_0) = \sum_{p_n \in \mathcal{R}} w(p_n) \cdot x(p_0 + p_n + \Delta p_n) \label{eq:dcn2}  
\end{align}
\end{subequations}

\subsection{Temporal processing}

The aforementioned CNN layers are connected to Long Short Term Memory (LSTM) \cite{hochreiter1997long} with a softmax layer to classify each frame as depicted in Fig. \ref{fig:cnn_lstm}. Including LSTM breaks the dependence on the size of network, because a much deeper CNN would be required to replace the RNN for video classification. Note that the dimension of the output space of LSTM is 8 only and there is no dropout as the focus is on comparing the CNN layers. 
The CNN processes every temporal slice of the video independently with the weights fixed.
In contrast, the LSTM has memory blocks to learn long term dependencies in the video since the output is dependent on the previous hidden state. 

As shown in Eq. (\ref{eq:lstm}), LSTM contains a cell state $c_t$, which is controlled by input gate $i_t$, forget gate $f_t$, and output gate $o_t$. $f_t$ decides which information to throw away, $i_t$ determines what to store in $c_t$, while $o_t$ generates the output based on the input $x_t$ as well as the previous hidden state $h_{t-1}$. The $W, b$ are the trainable parameters.  

\begin{subequations}
\label{eq:lstm}
\begin{align}
f_t = \sigma (W_f \cdot [h_{t-1}, x_t] + b_f) \label{eq:lstm1}  \\
i_t = \sigma (W_i \cdot [h_{t-1}, x_t] + b_i) \label{eq:lstm2} \\
g_t = \tanh (W_g \cdot [h_{t-1}, x_t] + b_g) \label{eq:lstm3}  \\
c_t = f_t \odot c_{t-1} + i_t \odot g_t \label{eq:lstm4}
\end{align}
\end{subequations}

\section{Dataset creation}

A practical challenge for training CNNs and RNNs for video classification with visual attention is the lack of datasets that exhibit sufficient variation in rotation and scale, because existing video classification benchmarks such as UCF101 \cite{soomro2012ucf101} and Sports-1M \cite{karpathy2014large} are collected from daily activities.
To address this issue, this work proposes a synthetic dataset based on Moving MNIST \cite{srivastava2015unsupervised}, with augmentation that includes rotation and scaling. This dataset is also overfit-resistant as a side benefit of applying augmentation, as analyzed in \cite{su2015render}. 

\begin{figure}[thpb]
  \centering
  \includegraphics[scale=0.48]{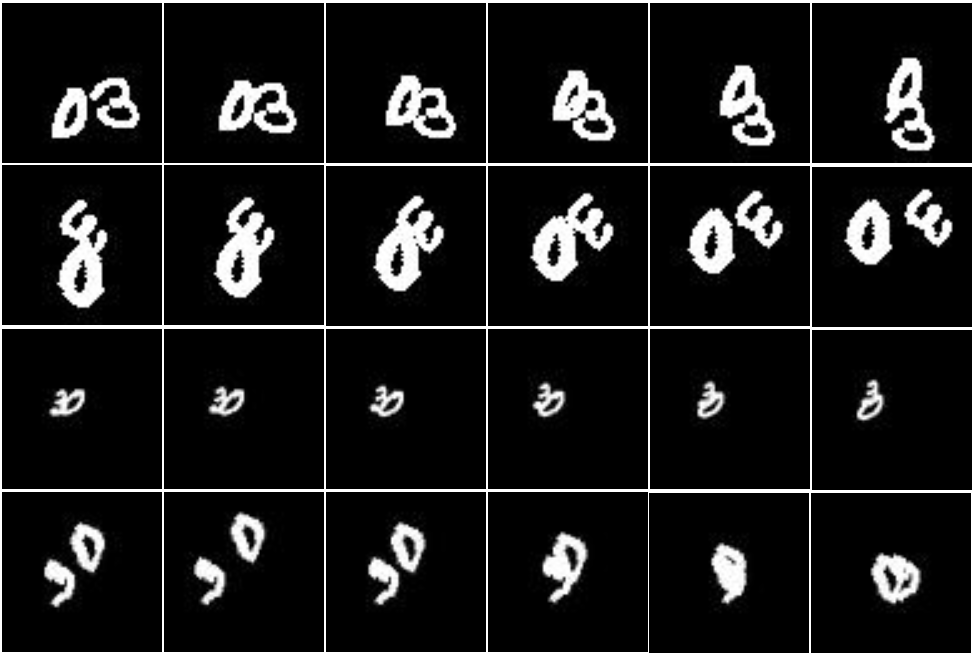}
  \caption{A sequence of images in class 03 in augmented Moving MNIST dataset. From first row to last row: original images, rotated images, scaled images, rotated and scaled images.}
  \label{fig:mnist}
\end{figure}

In our dataset, $D_{train}$ and $D_{test}$ that contain MNIST digits are created. Two samples $(\mathcal{I}_m, d_m), (\mathcal{I}_n, d_n) \in \mathcal{M}$ are randomly chosen from the MNIST training set, where $\mathcal{I}_m, \mathcal{I}_n$ are the images and $d_m, d_n \in \{0, 1, ..., 9\}$ are class labels. $\mathcal{I}_m, \mathcal{I}_n$ are placed at two random locations on a canvas $\mathcal{H}$ of size $64\times 64$. They are assigned with a velocity in range $[4, 12]$ whose direction is chosen at random. $\mathcal{I}_m, \mathcal{I}_n$ bounce off the boundary and could overlap. 
Furthermore, each canvas $\mathcal{H}$ is augmented with controllable variations of rotation and scaling to simulate the scene that contains multiple interacting objects, i.e. random rotation $\mathcal{R} \in [0, 2\pi]$ and scaling $\mathcal{S} \in [0.4, 1]$ are applied to obtain $\hat{\mathcal{H}}$. Several sequences of $\hat{\mathcal{H}}$ are used to produce the video $x_i$, and the labels for two digits are concatenated as the new class label $y_i = d_m d_n$. 
Sample images in the dataset are displayed in Fig. \ref{fig:mnist}.
Essentially, the video classification in this context is to recognize the digits in the presence of transformation and occlusion. 

\section{Experiment}

\subsection{Training details}

The networks are implemented bassed on \cite{stn, dcn, rcnn} using Keras \cite{chollet2015keras}. 
During pre-processing, the images are centered and converted to range $[0, 1]$. 
The models are pre-trained on the Scaled MNIST dataset, which is the MNIST dataset augmented by translation and scaling. The maximum random horizontal and vertical shifts in terms of fraction of total width and height is 0.2. Range for random zoom is $[1.0, 2.5]$. 
The kernel weights matrices are initialized via Xavier uniform initializer, and the bias terms are initialized to 0. The objective function is the cross entropy loss, and the optimizer for training is Adadelta optimizer with a learning rate of 1. 
The cross entropy loss and accuracy are $ 0.054, 98.31\%$ for LeNet, $0.026, 99.14\%$ for STN, and $0.072, 97.62\%$ for DCN. 

There are 100 classes in augmented Moving MNIST dataset, and 10 different rotations or scales in each class, with 5 frames for one setting. Hence there are 5000 images for training and 5000 images for testing. 
The training is conducted in 10 epochs with RMSProp optimizer using a learning rate of $1\times 10^{-3}$ and a batch size of 50. The objective function is categorical cross entropy loss. The kernel weights matrix in LSTM is initialized via Xavier uniform initializer, and the recurrent kernel weights matrix is initialized by a random orthogonal matrix. The bias vector is initialized to 0. 
Each time the LSTM is trained for 300 rounds and the loss and accuracy are averaged to reduce the variance. 

Note that the CNN and RNN are trained separately because STN and DCN introduce much more trainable parameters as shown in Fig. \ref{fig:cnn_lstm}. This issue is more severe for DCN as the number of trainable offsets increases linearly with respect to the size of the input feature map. For instance, there are 102,483,280 trainable parameters in total for DCN given a $64\times 64$ image, while only 6,431,080 parameters for LeNet, meaning that DCN has 16 times more parameters than LeNet.   
As a result, the $12$GB GPU memory used for the experiments becomes insufficient once the attention modules are distributed along the time dimension, making it difficult to train CNN and RNN together.  

\subsection{Quantitative results on augmented Moving MNIST}

\begin{table}[!ht]
\captionsetup{justification=centering}
 \caption {Comparison of cross entropy loss and test accuracy for the proposed model and baseline.} \label{tab:accuracy}
  \centering
	\begin{tabular}{c | c c c}
    \toprule
     Moving MNIST  & LeNet-LSTM & STN-LSTM & DCN-LSTM \\ \midrule
     Normal & $1.44, 97.96\%$ & $  1.98, 87.26\%$ & $1.27, 99.62\%$  \\ \midrule      
     Rotation & $ 1.42, 98.43\%$ & $ 1.97, 90.47\%$  & $ 1.29, 99.70\%$ \\ \midrule 
     Scaling & $ 1.52, 96.28\%$ & $ 1.99, 86.90\%$ & $  1.28, 99.41\%$ \\ \midrule    
    	 Rotation+Scaling& $1.51, 96.82\%$ & $ 1.99, 89.10\%$ & $1.25, 99.46\%$ \\
    \bottomrule
  \end{tabular}
\end{table}

The cross entropy loss and test accuracy are shown in Table \ref{tab:accuracy}. The proposed model STN-LSTM and DCN-LSTM are compared against LeNet-LSTM as baseline. 
As for video classification, DCN-LSTM consistently performs better than LeNet-LSTM, as the former has lower cross entropy loss and higher accuracy in all cases. This shows that although max-pooling layers could deal with rotation or scaling to some extent, DCN is superior than the max-pooling layers at handling deformations.  
Moreover, Table \ref{tab:accuracy} also shows that STN-LSTM performs poorly compared to LeNet-LSTM and DCN-LSTM. Since there are two digits in each image and STN performs affine transformation globally, it is hard to attend to the digits individually, resulting in the low accuracy.

Regarding different transformation, DCN-LSTM is hardly affected no matter which kind of augmentation is applied, as the accuracy stays above $99\%$. The variance in the accuracy of LeNet-LSTM is also small, indicating that max-pooling layers achieve similar results for different augmentation. In contrast, STN-LSTM handles rotated images significantly better than scaled images. 
Additionally, all models struggle with scaled images, as the accuracy in the third row is lowest. Furthermore, scaling is more challenging than rotation as the architectures could reach a relatively higher accuracy for rotated images.  

\subsection{Qualitative analysis on augmented Moving MNIST}

\begin{figure}[thpb]
  \centering
  \includegraphics[scale=0.45]{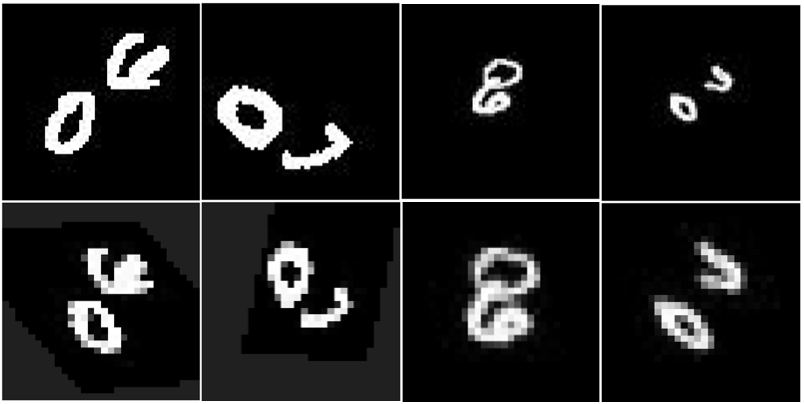}
  \caption{Output images of STN for class 06. First row: input images. Second row: output images of STN. From first column to last column: original images, rotated images, scaled images, rotated and scaled images.}
  \label{fig:stn_output}
\end{figure}

This section presents a qualitative analysis to shed some light on why STN does not do well. From the images processed by STN shown in Fig. \ref{fig:stn_output}, it could be observed that all the images are centered and transformed to a canonical orientation. There are several reasons when this kind of processing could be disadvantageous. For example, the original image is zoomed out when the digits are far from each other, even though doing so would make the digits smaller. This may be the reason why STN-LSTM has lower accuracy for images without any augmentation as shown in Table \ref{tab:accuracy}. 
In addition, STN rotates the two digits in the same way in the rotated image. This may be helpful when the same rotation is applied to both digits as what is done in the dataset, but not so when digits are rotated differently. 
To summarize, the inability of one STN module to attend to each digit separately affects its classification performance negatively. 
To address this issue, STN could be connected with RNN to detect multiple digits as proposed in \cite{sonderby2015recurrent}. The inference framework proposed in \cite{eslami1603attend} could be used as well to attend to scene elements one at a time.

\subsection{Classification of digit gesture}

\begin{figure}[thpb]
  \centering
  \includegraphics[scale=0.45]{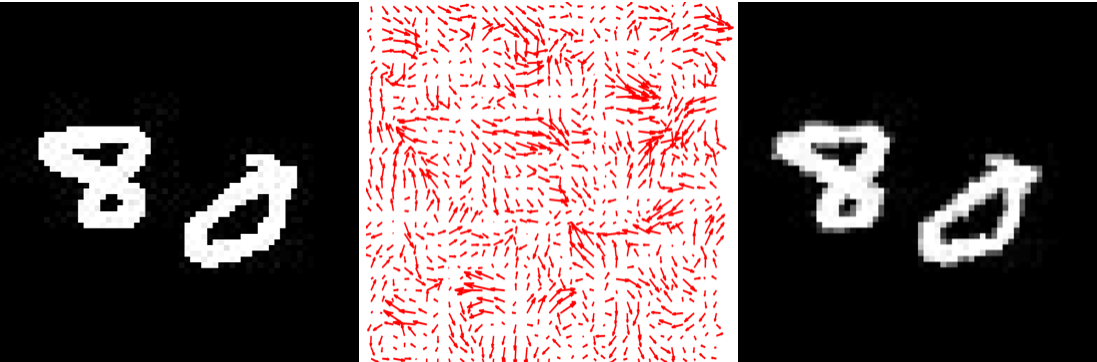}
  \caption{Image distorted by elastic deformation. From left to right: original image, elastic deformation field, deformed image.}
  \label{fig:deformed}
\end{figure}

Digit gesture has been widely used for evaluating the recognition algorithms, and classification of digit gesture is more challenging when the digits are written in the air by hand \cite{kim2017tracking}.
In this experiment, elastic deformation described in\cite{simard2003best} is applied to the images in Moving MNIST dataset, with rotation and scaling augmentation, to simulate oscillations of hand muscles. 
Random displacement fields $(\Delta x, \Delta y)$ are convolved with a Gaussian filter with standard deviation $\sigma = 10$, and size $[7, 7]$, then multiplied by a scaling factor $\alpha = 300$. 
The resulting images are similar to those in the TVC-hand gesture database \cite{kim2017tracking}, which contains images generated by accumulating the center points of a hand over time.
A sample image after applying elastic deformation is shown in Fig. \ref{fig:deformed}. 

The cross entropy loss and accuracy are $1.48, 97.19\%$ for LeNet-LSTM, $1.48, 97.19\%$ for STN-LSTM, and $1.28, 99.30\%$ for DCN-LSTM, indicating that DCN-LSTM is still the best model to deal with distorted digits because of its deformable convolutional layers.  
\cite{girshick2015deformable} shows that Deformable Parts Models (DPM) could be formulated as CNNs. 
Similarly, deformable convolutional layers in the DCN-LSTM could be modified to explicitly learn the deformation field, which is better than DPM for certain classes \cite{eth_biwi_01162}.   
Therefore, the capability of the proposed model in recognizing digit gesture suggests that the model could be extended for tracking hand trajectory or dealing with multi-part articulated objects such as hand.  

\section{Discussion}

This paper describes a spatiotemporal model for video classification with the presence of rotated and scaled objects.  The effectiveness of DCN-LSTM over the baseline is demonstrated using the augmented Moving MNIST dataset. 
Moreover, preliminary results on digit gesture recognition are also shown. 
Regarding future work, how to train the network end to end is worth exploring. It would be more elegant to train the two stages simultaneously in an efficient way.

\bibliographystyle{plainnat}
\bibliography{references}

\end{document}